\pdfoutput=1

\documentclass[11pt]{article}

\usepackage{emnlp2021}

\usepackage{times}
\usepackage{latexsym}

\usepackage[T1]{fontenc}

\usepackage[utf8]{inputenc}

\usepackage{microtype}

\usepackage{todonotes}
\usepackage{url}
\usepackage{amsmath, amsfonts}
\usepackage{multirow, multicol}
\usepackage{booktabs}
\usepackage{enumitem}
\usepackage{graphicx}
\usepackage{subcaption, caption}
\usepackage{xspace}
\usepackage{array}
\usepackage{inconsolata}
\newcolumntype{C}{>{$\boldmath$}c}

\usepackage{microtype}

\newcommand{\MSP}{\textit{MSP}\xspace}
\newcommand{\PPL}{\textit{PPL}\xspace}
\newcommand{\customie}{i.e.}

\newif\ifcomment
\commentfalse

\newcommand\sN{\ensuremath{\mathcal{N}}}

\newcommand\sX{\ensuremath{\mathcal{X}}}
\newcommand\sY{\ensuremath{\mathcal{Y}}}

\newcommand\BR{\ensuremath{\mathbb{R}}}

\newcommand\pc[1]{\ensuremath{\left\{ #1 \right\}}} 

\newcommand\refsec[1]{Section~\ref{sec:#1}}

\ifthenelse{\isundefined{\definition}}{\newtheorem{definition}{Definition}}{}
\ifthenelse{\isundefined{\assumption}}{\newtheorem{assumption}{Assumption}}{}
\ifthenelse{\isundefined{\hypothesis}}{\newtheorem{hypothesis}{Hypothesis}}{}
\ifthenelse{\isundefined{\proposition}}{\newtheorem{proposition}{Proposition}}{}
\ifthenelse{\isundefined{\theorem}}{\newtheorem{theorem}{Theorem}}{}
\ifthenelse{\isundefined{\lemma}}{\newtheorem{lemma}{Lemma}}{}
\ifthenelse{\isundefined{\corollary}}{\newtheorem{corollary}{Corollary}}{}
\ifthenelse{\isundefined{\alg}}{\newtheorem{alg}{Algorithm}}{}
\ifthenelse{\isundefined{\example}}{\newtheorem{example}{Example}}{}

\newcommand\hh[1]{\textcolor{blue}{[HH: #1]}}
\newcommand\hhdone[1]{\textcolor{teal}{[HH: #1]}}
\renewcommand\hh[1]{}
\renewcommand\hhdone[1]{}

\title{Types of Out-of-Distribution Texts and How to Detect Them}

\author{Udit Arora$^{\spadesuit}$ ~~
  William Huang$^{\clubsuit}$\thanks{\ \ Work done while at New York University.} ~~
  He He$^{\spadesuit}$ \\
  $^{\spadesuit}$New York University \\
  $^{\clubsuit}$Capital One \\
  \texttt{\{uditarora,hhe\}@nyu.edu, william.huang@capitalone.com}
}

\date{}

\begin{document}
\maketitle
\begin{abstract}

Despite agreement on the importance of detecting out-of-distribution (OOD) examples, there is little consensus on the formal definition of OOD examples and how to best detect them. We categorize these examples by whether they exhibit a \emph{background} shift or a \emph{semantic} shift, and find that the two major approaches to OOD detection, model calibration and density estimation (language modeling for text), have distinct behavior on these types of OOD data.
Across 14 pairs of in-distribution and OOD English natural language understanding datasets, we find that density estimation methods consistently beat calibration methods in {background} shift settings, while performing worse in {semantic} shift settings. 
In addition, we find that both methods generally fail to detect examples from challenge data, highlighting a weak spot for current methods.
Since no single method works well across all settings, our results call for an explicit definition of OOD examples when evaluating different detection methods. 

\end{abstract}

\section{Introduction}
Current NLP models work well when the training and test distributions are the same (e.g.\ from the same benchmark dataset).
However, it is common to encounter out-of-distribution (OOD) examples that diverge from the training data
once the model is deployed to real settings. 
When training and test distributions differ,
current models tend to produce unreliable or even catastrophic predictions that hurt user trust \citep{checklist:acl20}.
Therefore, it is important to identify OOD inputs 
so that we can modify models' inference-time behavior by abstaining, asking for human feedback, or gathering additional information \cite{DBLP:journals/corr/AmodeiOSCSM16}.


\begin{figure}
    \includegraphics[width=\columnwidth]{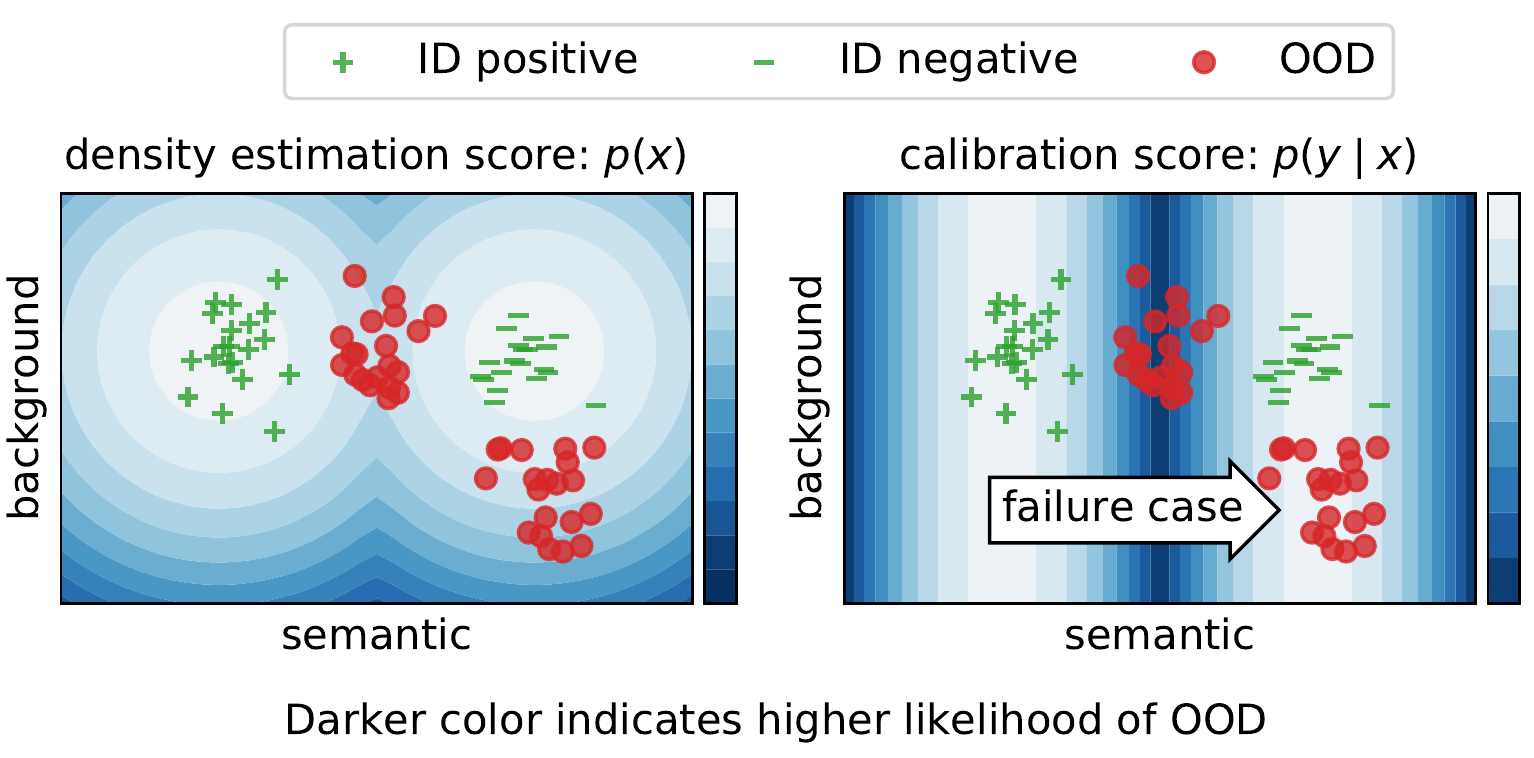}
    \caption{Illustration of semantic shift and background shift in $\BR^2$.
    Each point consists of semantic features ($x$-axis) and background features ($y$-axis).
    OOD examples (red points) can shift in either direction.
    The background color indicates regions of ID (light) and OOD (dark) given by
    the density estimation method (left) and the calibration method (right).
    The calibration method fails to detect OOD examples due to background shift.}
    \label{fig:OOD_example}
\end{figure}

Current work in NLP either focuses on specific tasks like intent classification in task-oriented dialogue \cite{ZhengCH20}, or arbitrary in-distribution (ID) and OOD dataset pairs \cite{hendrycks2020pretrained, DBLP:conf/iclr/HendrycksMD19, DBLP:journals/corr/abs-2104-08812}, e.g.\ taking a sentiment classification dataset as ID and a natural language inference dataset as OOD. However, getting inputs intended for a different task is rare in realistic settings as users typically know the intended task.
In practice, an example is considered OOD due to various reasons, 
e.g.\ being rare \cite{sagawa2020distributionally}, out-of-domain \cite{daume07easyadapt}, or adversarial \cite{carlini2017adversarial}.
This broad range of distribution shifts makes it unreasonable to expect a detection algorithm to work well for arbitrary OOD examples without assumptions on the test distribution \cite{DBLP:conf/aaai/AhmedC20}.

In this paper, we categorize OOD examples by common types of distribution shifts in NLP problems inspired by \citet{NEURIPS2019_1e795968} and \citet{DBLP:conf/cvpr/HsuSJK20}.
Specifically, we assume an input (e.g. a movie review)
can be represented as background features (e.g. genre)
that are invariant across different labels,
and semantic features (e.g. sentiment words) that are discriminative for the prediction task.
Correspondingly, at test time we consider two types of OOD examples characterized by a major shift in the distribution of background and semantic features, respectively.
While the two types of shifts often happen simultaneously, we note that there are realistic settings where distribution shift is dominated by one or the other. For example,
background shift dominates when the domain or the style of the text changes \cite{pavlick2017style},
e.g. from news to tweets,
and semantic shift dominates when unseen classes occur at test time, as in open-set classification \cite{scheirer2013toward}.\footnote{
    We exclude \emph{task} shift where the OOD examples are from a different task,
    e.g. textual entailment inputs for a text classification model,
    because it is less likely to happen in realistic settings where users are often aware of the intended use of the model.
}

We use this categorization to evaluate two major approaches to OOD detection,
namely calibration methods that use the model's prediction confidence \citep{hendrycks2016baseline, DBLP:conf/iclr/LiangLS18}
and density estimation methods that fit a distribution of the training inputs \citep{DBLP:conf/iclr/NalisnickMTGL19, DBLP:journals/corr/abs-2007-05566, DBLP:conf/nips/KirichenkoIW20}.
We show that the two approaches make implicit assumptions on the type of distribution shift, and
result in behavioral differences under each type of shift.
%
%
By studying ID/OOD pairs constructed from both simulations and real datasets,
we find that the density estimation method better accounts for shifts in background features, consistently outperforming the calibration method on \textit{background} shift pairs.
We further see the opposite in \textit{semantic} shift pairs, with the calibration method consistently yielding higher performance.

In addition, we analyze the detection performance on challenge datasets \cite{mccoy2019hans,naik2018stress}
through the lens of background/semantic shift.
We find that these challenge datasets provide interesting failure cases for both methods.
Calibration methods completely fail when the model is over-confident due to spurious semantic features.
While density estimation methods are slightly more robust,
language models are easily fooled by repetitions that significantly increase the probability of a piece of text.
Together, our findings suggest that better definitions of OOD and corresponding evaluation datasets are required
for both model development and fair comparison of OOD detection methods.


\section{Categorization of OOD Examples}
\label{sec:ood-text}
\subsection{Problem Statement}
Consider classification tasks where each example consists of an input $x\in\sX$ and its label $y\in\sY$.
In the task of OOD detection, we are given a training dataset $\mathcal{D}_{\text{train}}$ of $(x,y)$ pairs sampled from the training data distribution $p(x,y)$.
At inference time, given an input $x'\in\sX$ the goal of OOD detection is to identify whether $x^\prime$ is a sample drawn from $p(x,y)$.

\subsection{Types of Distribution Shifts}
\label{subsec:types_ood}
As in \citep{NEURIPS2019_1e795968}, we assume that any representation of the input $x$, $\phi(x)$, can be decomposed into two independent and disjoint components:
the background features $\phi_b(x)\in\BR^m$ and the semantic features $\phi_s(x)\in\BR^n$. Formally, we have
\begin{align}
    \phi(x) &= [\phi_s(x); \phi_b(x)],\\
    p(x) &= p(\phi_s(x))p(\phi_b(x))
\end{align}
Further, we assume that $\phi_b(x)$ is independent of the label while $\phi_s(x)$ is not. Formally, $\forall y\in\sY$,
\begin{align}
p(\phi_b(x) \mid y) = p(\phi_b(x)), \\
p(\phi_s(x) \mid y) \neq p(\phi_s(x))
\end{align}
\textcolor{black}{Note that $p$ refers to the ground truth distribution, as opposed to one learned by a model.}

Intuitively, the background features consist of population-level statistics that do not depend on the label, whereas the semantic features have a strong correlation with the label. \textcolor{black}{A similar decomposition is also used in previous work on style transfer \citep{DBLP:conf/aaai/FuTPZY18}, where a sentence is decomposed into the content (semantic) and style (background) representations in the embedding space.} 

Based on this decomposition, we classify the types of OOD data as either \textit{semantic} or \textit{background} shift based on whether the distribution shift is driven by changes in $\phi_s(x)$ or $\phi_b(x)$, respectively. 
An example of background shift is a sentiment classification corpus with reviews from IMDB versus GoodReads where phrases indicating positive reviews (e.g. ``best'', ``beautifully'') are roughly the same while the background phrases change significantly (e.g. ``movie'' vs ``book'').
On the other hand, semantic shift happens when we encounter unseen classes at test time, e.g.\ a dialogue system for booking flight tickets receiving a request for meal vouchers \cite{ZhengCH20}, \textcolor{black}{or a question-answering system handling unanswerable questions \cite{rajpurkar2018squadrun}.} 
We note that the two types of shifts may happen simultaneously in the real world,
and our categorization is based on the most prominent type of shift. 

\section{OOD Detection Methods}

To classify an input $x\in\sX$ as ID or OOD, we produce a score $s(x)$ and classify it as OOD if $s(x) < \gamma$,
where $\gamma$ is a pre-defined threshold.
Most methods differ by how they define $s(x)$.
Below we describe two types of methods commonly used for OOD detection.

\paragraph{Calibration methods.}
These methods use the model's prediction confidence as the score.
A well-calibrated model's confidence score reflects the likelihood of the predicted label being correct. Since the performance on OOD data is usually lower than on ID data,
lower confidence suggests that the input is more likely to be OOD.
The simplest method to obtain the confidence score is to directly use
the conditional probability produced by a probabilistic classifier $p_{\text{model}}$, referred to as maximum softmax probability  \cite[\MSP;][]{hendrycks2016baseline}.
Formally,
\begin{align}
\label{eqn:msp}
    s_{\MSP}(x) = \max_{k\in\sY} p_\text{model}(y=k\mid x).
\end{align}
While there exist more sophisticated methods that take additional calibration steps \cite{DBLP:conf/iclr/LiangLS18, lee2018simple},
\MSP proves to be a strong baseline, especially when $p_\text{model}$ is fine-tuned from pre-trained transformers \cite{hendrycks2020pretrained, desai2020calibration}.

\paragraph{Density estimation methods.}
These methods use the likelihood of the input given by a density estimator as the score. 
For text or sequence data, a language model $p_\text{LM}$ is typically used to estimate $p(x)$ \cite{NEURIPS2019_1e795968}.
To avoid bias due to the length of the sequence (see analysis in Appendix \ref{app:prob}),
we use the token perplexity (\PPL) as the score. 
Formally, given a sequence $x=(x_1, \ldots, x_T)$,
\begin{align}
\label{eqn:ppl}
    s_{\PPL}(x) = \exp\pc{\frac{1}{T} \sum_{t=1}^T \log p_{\text{LM}}(x_t\mid x_{1:t-1}) }
\end{align}
While there are many works on density estimation methods using flow-based models in computer vision \citep[e.g.][]{DBLP:conf/iclr/NalisnickMTGL19,zhang2020open}, there is limited work experimenting with density estimation methods for OOD detection on text \cite{DBLP:journals/corr/abs-2006-04666}.

\paragraph{Implicit assumptions on OOD.}
One key question in OOD detection is how the distribution shifts at test time,
i.e.\ what characterizes the difference between ID and OOD examples.
Without access to OOD data during training, the knowledge must be incorporated into the detector through some inductive bias.
Calibration methods rely on $p(y\mid x)$ estimated by a classifier,
thus they are more influenced by the semantic features which are correlated with the label.
We can see this formally by
\begin{align}
    p(y\mid x) &\propto p(x\mid y)p(y) \\
    &= p(\phi_b(x) \mid y) p(\phi_s(x) \mid y) p(y)\\
    &\propto p(\phi_s(x) \mid y) p(y)
    .
\end{align}
In contrast, density estimation methods are sensitive to all components of the input, including both background and semantic features,
even in situations where distribution shifts are predominately driven by one particular type.
In the following sections, we examine how these implicit assumptions impact performance on different ID/OOD pairs.




\section{Simulation of Distribution Shifts}
\label{sec:toy}


As an illustrative example, we construct a toy OOD detection problem using a binary classification setting similar to the one depicted in Figure \ref{fig:OOD_example}. This allows us to remove estimation errors and study optimal calibration and density estimation detectors under controlled semantic and background shifts.

\subsection{Data Generation}
\label{subsec:toy_generation}
We generate the ID examples from a Gaussian Mixture Model (GMM):
\begin{align}
    y &= \begin{cases}
        0 & \text{w.p. } 0.5 \\
        1 & \text{otherwise}
    \end{cases}, \\
    x \mid y=i &\sim \sN(\mu^i, \Sigma) .
\end{align}
The centroids are sets of semantic and background features such that
$\mu^1=[\mu_s, \mu_b]$ and
$\mu^0=[-\mu_s, \mu_b]$,
where $\mu_s\in \BR^n$ and $\mu_b\in \BR^m$.
In the 2D case in Figure \ref{fig:OOD_example}, this corresponds to the two Gaussian clusters where the first component is the semantic feature and the second is the background feature.


In this case, we know the true calibrated score $p(y\mid x)$
and the true density $p(x)$ given any inputs.
Specifically, the optimal classifier is given by the Linear Discriminant Analysis (LDA) predictor.
By setting $\Sigma$ to the identity matrix,
it corresponds to a linear classifier with weights $[2\mu_s, \mathbf{0}_b]$, where $\mathbf{0}_b \in \BR^m$ is a vector of all $0$s.
For simplicity, we set $\mu_s = \mathbf{1}_s$ and $\mu_b = \mathbf{0}_b$, where $\mathbf{1}_s \in \BR^n, \mathbf{0}_b \in \BR^m$ are vectors of all $0$s.

\subsection{Semantic Shift}
We generate sets of OOD examples using a semantic shift by varying the overlap of ID and OOD semantic features. Formally, we vary the overlap rate $r$ such that
\begin{align}
    r &= \frac{\vert \mu_s \cap \mu_s^{\text{Shift}} \vert}{\vert \mu_s \vert}
\end{align}
where $\mu_s, \mu_s^{\text{Shift}} \in \BR^n$ are the set of semantic features for ID and OOD, respectively, $\mu_s \cap \mu_s^{\text{Shift}}$ represents the common features between the two, and $\vert \cdot \vert$ denotes the number of elements.

We fix the total dimensions to $n + m = 200$ and set $n = 40$ (semantic features) and $m = 160$ (background features). Further, we vary $r$ by increments of $10\%$. Larger $r$ indicates stronger semantic shift. For each $r$, we randomly sample ID and OOD semantic features and report the mean over $20$ trials with $95\%$ confidence bands in Figure \ref{fig:gaussian_toy}.

\subsection{Background Shift}

We generate sets of OOD examples using a background shift by applying a displacement vector $z=[\mathbf{0}_s, z_b]$ to the two means. Formally,
\begin{align}
    \mu^{i, \text{ Shift}} = \mu^i + z
\end{align}
where $\mathbf{0}_s  \in \BR^n$ is a vector of all $0$s.

We set $z = \alpha [\mathbf{0}_s, \mathbf{1}_b]$, where $\mathbf{1}_b \in \BR^m$ is a vector of $1$s. Note that this shift corresponds to a translation of the ID distribution along the direction of $\mu_b$. We set the total dimensions to $n + m = 200$ while varying the split between semantic ($n$) and background ($m$) components by increments of $20$.

\subsection{Simulation Results}

\begin{figure}[h]
    \centering
    \includegraphics[width=0.5\textwidth]{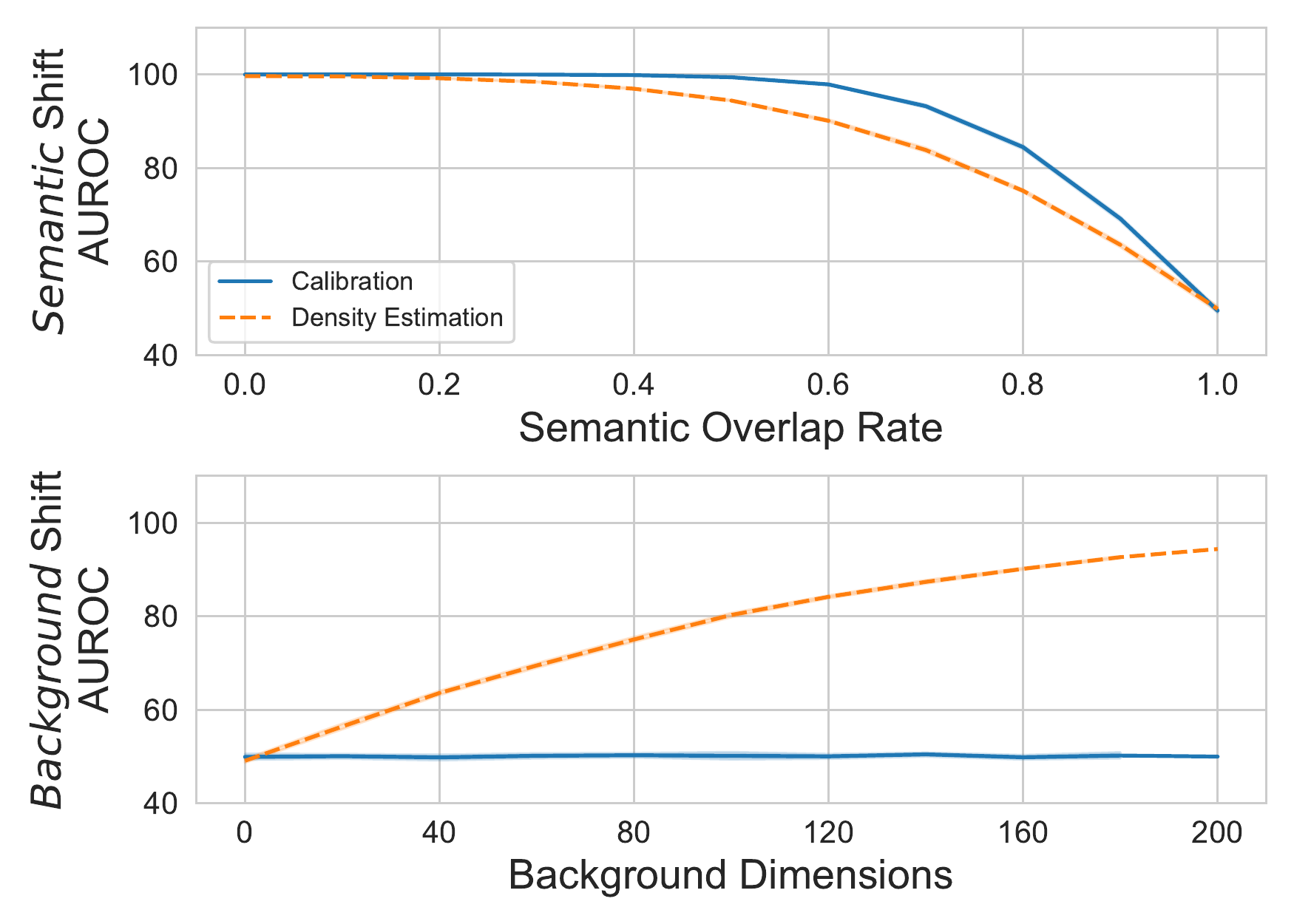}
    \caption{Area Under Receiver Operating Characteristics (AUROC) of calibration (blue) and density estimation (orange) methods for OOD detection using our toy binary classification problem. The calibration method outperforms the density estimation method under larger semantic shifts while the opposite is true under larger background shifts.}
    \label{fig:gaussian_toy}
\end{figure}

Figure \ref{fig:gaussian_toy} shows the OOD detection performance of our simulated experiment. We use Area Under the Receiver Operating Characteristics (AUROC) as our performance metric. 


We see that the calibration method generally outperforms density estimation. Further, the performance gap between the two methods decreases as both methods approach near-perfect performance under large semantic shifts with no overlap in semantic features,
and approach chance under no semantic shift with completely overlapping semantic features. However, the calibration method is unable to improve performance under background shifts in either regime because the background features do not contribute to $p(y\mid x)$ as the LDA weights are $0$ for these components (Section \ref{subsec:toy_generation}). We find these results in line with our expectations and use them to drive our intuition when evaluating both types of OOD detection methods for real text data.

\section{Experiments and Analysis}
\label{sec:experiments}
We perform head-to-head comparisons of calibration and density estimation methods on 14 ID/OOD pairs categorized as either background shift or semantic shift, as well as 8 pairs from challenge datasets.

\subsection{Setup}
\paragraph{OOD detectors.}
Recall that the \textbf{calibration method}~ \MSP relies on a classifier trained on the ID data.
We fine-tune the RoBERTa \citep{DBLP:journals/corr/abs-1907-11692} model on the ID data and compute its prediction probabilities (see Equation \eqref{eqn:msp}). 
For the \textbf{density estimation} method \PPL,
we fine-tune GPT-2 \citep{radford2019language} on the ID data
and use perplexity as the OOD score (see Equation \eqref{eqn:ppl}).\footnote{
We also use the sentence probability ($p(x)$) as the score, but find it highly sensitive to sequence lengths (Appendix \ref{app:prob}).
} 
 %
 To control for model size of the two methods, we choose RoBERTa$_{\rm{Base}}$ and GPT-2$_{\rm{Small}}$, which have 110M and 117M parameters, respectively. We also experiment with two larger models, RoBERTa$_{\rm{Large}}$ and GPT-2$_{\rm{Medium}}$ with 355M and 345M parameters, respectively.
 
 We evaluate the OOD detectors by AUROC and the False Alarm Rate at 95\% Recall (FAR95), which measures the misclassification rate of ID examples at 95\% OOD recall. Both metrics show similar trends (see Appendix \ref{app:far95} for FAR95 results).
 
 \paragraph{Training details.}
For RoBERTa, we fine-tune the model for 3 epochs on the training split of ID data with a learning rate of 1e-5 and a batch size of 16. For GPT-2, we fine-tune the model for 1 epoch on the training split of ID data for the language modeling task, using a learning rate of 5e-5 and a batch size of 8. \footnote{Our code can be found at \url{https://github.com/uditarora/ood-text-emnlp}.}

\paragraph{Oracle detectors.}
To get an estimate of the upper bound of OOD detection performance, we consider the situation where we have access to the OOD data and can directly learn an OOD classifier. Specifically, we train a logistic regression model with bag-of-words features using 80\% of the test data and report results on the remaining 20\%.



\subsection{Semantic Shift}
Recall that the distribution of discriminative features changes
in the semantic shift setting, i.e.
$p_{\text{train}}(\phi_s(x)) \neq p_{\text{test}}(\phi_s(x))$
(\refsec{ood-text}).
We create semantic shift pairs by including test examples from classes unseen during training.
Thus, semantic features useful for classifying the training data are not representative in the test set.

We use the News Category \citep{news_category_dataset} and DBPedia Ontology Classification \citep{NIPS2015_250cf8b5} multiclass classification datasets to create two ID/OOD pairs. The News Category dataset consists of HuffPost news data. We use the examples from the five most frequent classes as ID (News Top-5) and the data from the remaining 36 classes as OOD (News Rest). The DBPedia Ontology Classification dataset consists of data from Wikipedia extracted from 14 non-overlapping classes of DBPedia 2014 \citep{lehmann2015dbpedia}. We use examples from the first four classes by class number as ID (DBPedia Top-4) and the rest as OOD (DBPedia Rest).


\paragraph{Results.}

\begin{table}
    \small
    \centering
    \resizebox{\columnwidth}{!}{%
    \begin{tabular}{llllr}
        \toprule
        \multirow{2}{*}{ID} &  \multirow{2}{*}{OOD} & \multicolumn{3}{c}{AUROC} \\
        & & \PPL & \MSP & Oracle \\
        \midrule
        News Top-5 & News Rest & 60.2 & \textbf{78.9} & 72.0 \\
        DBPedia Top-4 & DBPedia Rest & 75.4 & \textbf{88.8} & 99.6 \\
        \bottomrule
    \end{tabular}
    }
    \caption{Performance on semantic shifts, with higher score (among \PPL/\MSP) in \textbf{bold}. We can see that the calibration method using \MSP~significantly outperforms the density estimation methods.}
    \label{tab:semantic-shift}
\end{table}

Table \ref{tab:semantic-shift} shows the results for our semantic shift pairs. The calibration method consistently outperforms the density estimation method, indicating that calibration methods are better suited for scenarios with large semantic shifts, which is in line with our simulation results (\refsec{toy}).

\hh{The takeaway should probably focus on semantic shift, e.g. they are good at detecting semantic shift. Need comments on PPL method too: it still works, but is probably influenced by the large amounts of background features (give examples). Would be nice to refer to relevant simulation results too.} 

\subsection{Background Shift}
Recall that background features (e.g. formality) do not depend on the label.
Therefore, we consider domain shift in sentiment classification and natural language inference (NLI) datasets.

For our analysis, we use the SST-2 \cite{socher2013SST2}, IMDB \citep{maas2011IMDB}, and Yelp Polarity \citep{NIPS2015_250cf8b5} binary sentiment classification datasets. The SST-2 and IMDB datasets consist of movie reviews with different lengths. Meanwhile, the Yelp polarity dataset contains reviews for different businesses, representing a domain shift from SST-2 and IMDB. Each of these datasets is used as ID/OOD, using the validation split of SST-2 and test split of IMDB and Yelp Polarity for evaluation.

We also use the SNLI \cite{DBLP:conf/emnlp/BowmanAPM15}, MNLI \cite{N18-1101} and RTE (from GLUE, \citeauthor{wang18GLUE}, \citeyear{wang2018glue}) datasets. SNLI and MNLI consist of NLI examples sourced from different genres. RTE comprises of examples sourced from a different domain.
Where there is some change in semantic information since the task has the two labels (\textit{entailment} and \textit{non-entailment}) as opposed to three (\textit{entailment}, \textit{neutral} and \textit{contradiction}) in SNLI and MNLI,\footnote{Both neutral and contradiction are considered as non-entailment when evaluating accuracy with RTE vs SNLI/MNLI or vice-versa.}
domain/background shift is more prominent since the semantic features for the NLI task are similar. Each of these datasets is used as either ID or OOD, and we use the validation set of the OOD data for evaluation.

\paragraph{Results.}


\begin{table}
    \centering
    \small
    \resizebox{\columnwidth}{!}{%
    \begin{tabular}{ll>{\bfseries}rrrrr}
        \toprule
        \multirow{2}{*}{ID} &  \multirow{2}{*}{OOD} & \multicolumn{3}{c}{AUROC} & \multicolumn{2}{c}{Accuracy} \\
        & & \textnormal{\PPL} & \MSP & Oracle & OOD ($\Delta$) & ID \\
        \midrule
        \multirow{2}{*}{SST-2} 
             &       IMDB & 97.9 & 66.2 &   100.0 & 92.0 (-1.8) & \multirow{2}{*}{93.8} \\
             &       Yelp & 98.7 & 57.5 &    99.8 & 94.4 (+0.6) & \\
        \midrule
        \multirow{2}{*}{IMDB} &      SST-2 & 96.9 & 82.6 &   100.0 & 89.2 (-6.3) & \multirow{2}{*}{95.5} \\
            &       Yelp & 77.9 & 67.1 &   100.0 & 95.4 (-0.1) & \\
        \midrule
        \multirow{2}{*}{Yelp} &      SST-2 & 98.9 & 85.9 &    99.8 & 88.9 (-9.3) & \multirow{2}{*}{98.2} \\
            &       IMDB & 86.6 & 61.8 &   100.0 & 93.2 (-5.0) & \\
        \midrule
        \multirow{2}{*}{SNLI} 
             &       RTE & 94.6 & 78.7 &   99.8 & 67.5 (-22.6) & \multirow{2}{*}{90.1} \\
             &       MNLI & 96.7 & 75.6 &    99.7 & 77.9 (-12.2) & \\
        \midrule
        \multirow{2}{*}{RTE} &      SNLI & 81.2 & 45.1 &   99.7 & 82.0 (+6.9) & \multirow{2}{*}{75.1} \\
            &       MNLI & 81.4 & 55.5 &   97.0 & 77.3 (+2.2) & \\
        \midrule
        \multirow{2}{*}{MNLI} &      SNLI & 75.7 & 56.1 &    99.7 & 80.4 (-4.4) & \multirow{2}{*}{84.8} \\
            &       RTE & \textnormal{68.0} & \textbf{76.5} &   96.7 & 76.5 (-8.3) & \\
        \bottomrule
    \end{tabular}
    }
    \caption{Performance on background shifts caused by shift in domain. For each pair, higher score obtained (by \PPL~or \MSP) is in \textbf{bold}. The density estimation method using \PPL~outperforms the calibration method.}
    \label{tab:domain-shift}
\end{table}


Table \ref{tab:domain-shift} shows the results for binary sentiment classification and NLI domain shifts. The density estimation method consistently outperforms the calibration method (for all pairs except MNLI vs RTE), indicating that \PPL~ is more sensitive to changes in background features. Further, in cases where the discriminative model generalizes well (as evident by the small difference in ID and OOD accuracy numbers), we find that the calibration method performance is close to random (50) because a well-calibrated model also has higher confidence on its correct OOD predictions.

We note that the discriminative models tend to generalize well here, hence it might be better to focus on domain adaptation instead of OOD detection when the shift is predominantly a background shift. We discuss this further in Section \ref{sec:related}. 


\subsection{Analysis}
\paragraph{Controlled distribution shifts.} We use two controlled distribution shift experiments on real text data to further study the framework of semantic and background shifts.
For background shift, we append different amounts of text from Wikitext \citep{DBLP:conf/iclr/MerityX0S17} and Civil Comments \citep{borkan2019CivilComments} to SST-2 examples to create synthetic ID and OOD examples, respectively. We append the unrelated texts with lengths $\in (25,50,100,150,200)$ words.
For semantic shift, we use the News Category dataset and move classes from ID to OOD. We start with the top 40 ID classes by frequency and move classes in increments of 10. The ID coverage of semantic information decreases as more classes move to the OOD subset, resulting in a larger semantic shift.

\paragraph{Results.}
\begin{figure}[h]
    \centering
    \begin{subfigure}[]{0.5\textwidth}
        \centering
        \includegraphics[width=\textwidth]{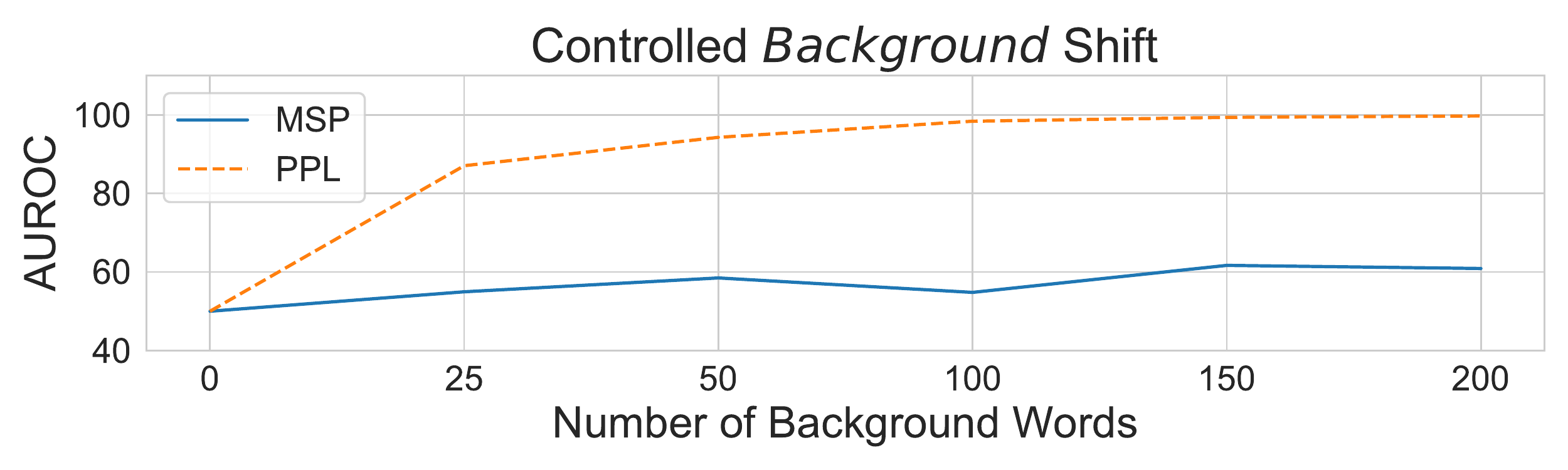}
        \label{subfig:synthetic-background}
    \end{subfigure}
    
    \begin{subfigure}[]{0.5\textwidth}
        \includegraphics[width=\textwidth]{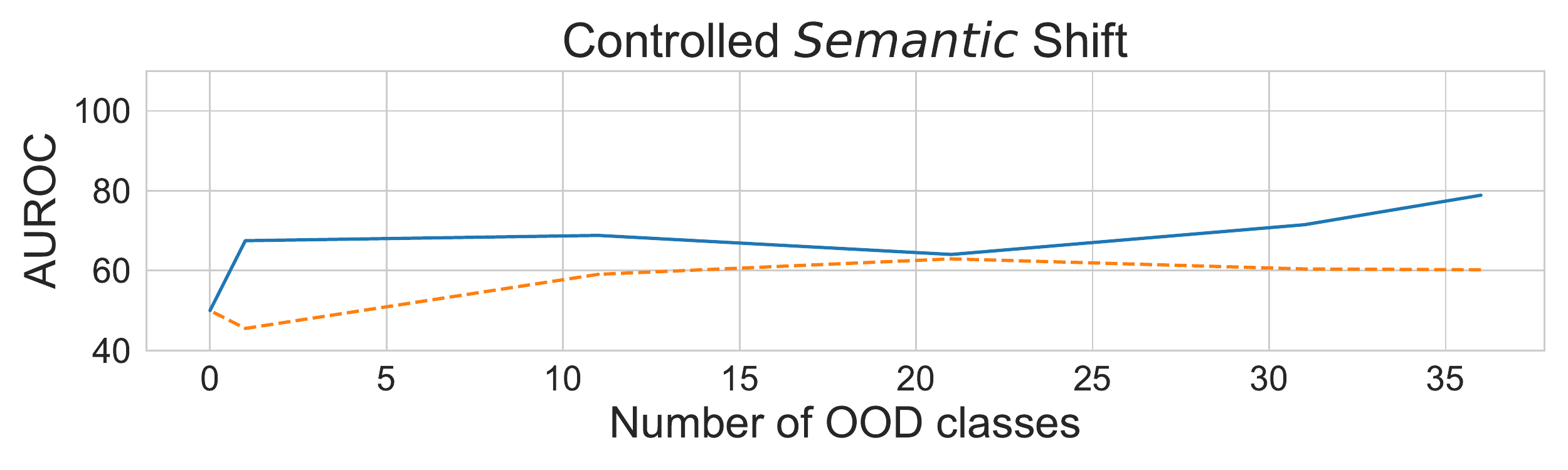}
        \label{subfig:synthetic-semantic}
    \end{subfigure}
    \caption{AUROC of \PPL~(orange) and \MSP~(blue) for controlled background and semantic shift experiments. The density estimation method performance improves as we increase the amount of background shift by appending longer texts, and the calibration method performance increases as we increase the amount of semantic shift by moving more classes to OOD.}
    \label{fig:synthetic-experiments}
\end{figure}

Figure \ref{fig:synthetic-experiments} shows the AUROC score obtained from both methods for our controlled distribution shift experiments. We see that the density estimation method is more sensitive to the amount of synthetic background text than calibration methods, and that the calibration method is more sensitive to the number of ID/OOD classes. This is in line with our intuition about the shifts and the results we obtain from simulated data (\refsec{toy}).




\begin{table}
    \centering
    \small
    \setlength\tabcolsep{3pt}
    \resizebox{\columnwidth}{!}{%
    \begin{tabular}{llrrrrr}
    \toprule
    \multirow{2}{*}{ID} &  \multirow{2}{*}{OOD} & \multicolumn{2}{c}{Base} & \multicolumn{2}{c}{Large} & \\
        & & \PPL & \MSP & \PPL & \MSP & Oracle \\
    \midrule
         IMDB &       Yelp & \textbf{77.9} & 67.1 &       75.5 &       74.5 & 100.0 \\
         News Top-5 &  News Rest & 60.2 & 78.9 &       61.7 &       \textbf{79.1} & 72.0 \\
    \bottomrule
    \end{tabular}
    }
    \caption{Performance of Base and Large models for a background shift pair and semantic shift pair each, with higher score in \textbf{bold}. The larger discriminative model helps close the performance gap between the calibration method and density estimation method for background shift.}
    \label{tab:large-performance}
\end{table}

\paragraph{Larger models.}
Table \ref{tab:large-performance} shows the results using larger models for OOD detection. We observe that the larger discriminative model achieves a much higher score for the background shift pair, closing the gap with the language model performance. We speculate that the larger model is able to learn some of the background features in its representation. The performance for the semantic shift pair is largely unchanged when using the larger models.


\subsection{Challenge Data}
Challenge datasets are designed to target either superficial heuristics adopted by a model (e.g. premise-hypothesis overlap) or model deficiencies (e.g. numerical reasoning in NLI), which creates significant challenges for deployed models \cite{checklist:acl20}. It is therefore desirable to abstain on detected OOD examples. We consider the following challenge datasets. 

\paragraph{Human-generated challenge data.}
\citet{kaushik2020cIMDB} crowdsourced a set of counterfactually-augmented IMDB examples (c-IMDB) by instructing annotators to minimally edit examples to yield counterfactual labels.
\textcolor{black}{This changes the distribution of semantic features with high correlation to labels such that $p_\text{train}(\phi_s(x)) \neq p_\text{test}(\phi_s(x))$, creating a semantic shift. We consider IMDB as ID and c-IMDB as OOD, combining the training, validation, and test splits of c-IMDB for evaluation.}

\paragraph{Rule-based challenge data.}
HANS \citep{mccoy-etal-2019-right} consists of template-based examples that have high premise-hypothesis overlap but are non-entailment, which mainly results in background shift due to the specific templates/syntax. 
Similarly, the Stress Test dataset \citep{naik-etal-2018-stress} is a set of automatically generated examples designed to evaluate common errors from NLI models. We categorize the type of distribution shifts from these test categories with respect to MNLI (ID) depending on whether they append ``background'' phrases to the ID examples or replace discriminative phrases (Table \ref{tab:MNLI-challenge}).

\textcolor{black}{Antonym (changing premise to obtain an antonymous hypothesis resulting in contradiction despite high overlap) and Numerical Reasoning (different semantic information than MNLI training set) constitute semantic shifts, as the set of semantic features now focus on specific types of entailment reasoning (e.g. antonymy and numerical representation). Negation (appending ``and false is not true'' to hypothesis), Spelling Errors (randomly introducing spelling errors in one premise word), Word Overlap (appending ``and true is true'' to each hypothesis), and Length Mismatch (appending a repetitive phrase ``and true is true'' five times to the premise) constitute background shifts because they introduce population level changes (e.g. appending ``and true is true'' to each hypothesis) that are unrelated to the entailment conditions of each example.}

\textcolor{black}{We consider the matched Negation, Spelling Errors, Word Overlap and Length Mismatch examples from the Stress Test as background shifts, and the Numerical Reasoning and Antonym examples as semantic shifts. We consider MNLI as ID for these challenge examples and use the validation split of HANS and MNLI for evaluation.}

\begin{table}
    \small
    \centering
    \setlength\tabcolsep{3pt}
    \resizebox{\columnwidth}{!}{%
    \begin{tabular}{lllr>{\bfseries}rr}
    \toprule
    \multirow{2}{*}{ID} &  \multirow{2}{*}{OOD} & \multirow{2}{*}{Shift} & \multicolumn{3}{c}{AUROC} \\
        & & & \PPL & \textnormal{\MSP} & Oracle \\
    \midrule
    IMDB & c-IMDB & Semantic & 53.5 & \textbf{63.7} & 77.5 \\
    \midrule
     \multirow{7}{*}{MNLI} & HANS & Background & \textbf{98.3} & \textnormal{55.0} &   100.0 \\
      &            Negation & Background & 44.5 & 60.5 &    99.9 \\
      &     Len. Mismatch & Background & 19.6 & 51.6 &   100.0 \\
      &      Spell. Error & Background & 43.9 & 57.7 &    98.4 \\
      &        Word Overlap & Background & 42.4 & 61.7 &    99.8 \\
      & Antonym & Semantic &  4.5 & 55.3 &    97.3 \\
      & Num. Reason. & Semantic & 27.5 & 75.8 &    99.7 \\
    \bottomrule
    \end{tabular}
    }
    \caption{AUROC scores obtained using \PPL, \MSP~and Oracle for challenge data. The primary type of shift observed is indicated in the `Shift' column. Higher performance (among \MSP/\PPL) for each pair is in \textbf{bold}. We can see that both methods struggle with most types of challenge data.}
    \label{tab:MNLI-challenge}
\end{table}

\paragraph{Failure case 1: spurious semantic features.}
Challenge data is often constructed to target \emph{spurious features} (e.g. premise-hypothesis overlap for NLI) that are useful on the training set but do not correlate with the label in general, e.g. on the test set.
Therefore, a discriminative model would be \emph{over-confident} on the OOD examples because the spurious semantic features that were discriminative during training, while still prominent, are no longer predictive of the label.
As a result, in Table \ref{tab:MNLI-challenge}, \MSP struggles with most challenge data, achieving an AUROC score close to random (50). 
On the other hand, the density estimation method achieves almost perfect performance on HANS.

\paragraph{Failure case 2: small shifts.}
While density estimation methods perform better in background shift settings, our simulation results show that they still struggle to detect small shifts when the ID and OOD distributions largely overlap.
Table \ref{tab:MNLI-challenge} shows similar findings for Negation and Word Overlap Stress Test categories that append short phrases (e.g. ``and true is true'') to each ID hypothesis.

\paragraph{Failure case 3: repetition.}
For Antonym, Numerical Reasoning, and Length Mismatch, \PPL performance is \emph{significantly worse than random}, indicating that our language model assigns higher likelihoods to OOD than ID examples.
These challenge examples contain highly repetitive phrases
(e.g. appending ``\textit{and true is true}'' five times in Length Mismatch, or high overlap between premise and hypothesis in Numerical Reasoning and Antonym), which is known to yield high likelihood under recursive language models \cite{holtzman2019curious}.
Thus repetition may be used as an attack to language model-based OOD detectors.

Overall, the performance of both methods drops significantly on the challenge datasets.
Among these, human-generated counterfactual data is the most difficult to detect,
and rule-based challenge data can contain unnatural patterns that cause unexpected behavior.

\subsection{Discussion}
The performance of calibration and density estimation methods on OOD examples categorized along the lines of semantic and background shift provides us with insights that can be useful in improving OOD detection. This framework can be used to build better evaluation benchmarks that focus on different challenges in OOD detection. A choice between the two methods can also be made based on the anticipated distribution shift at test time, i.e, using calibration methods when detecting semantic shift is more important, and using density estimation methods to detect background shifts.
However, we observe failure cases from challenge examples, with density estimation methods failing to detect OOD examples with repetition and small shifts, and calibration methods failing to detect most challenge examples. This indicates that these challenge examples constitute a type of OOD that target the weaknesses of both approaches. This highlights the room for a more explicit definition of OOD to progress the development of OOD detection methods and create benchmarks that reflect realistic distribution shifts.

\section{Related Work}
\label{sec:related}

\paragraph{Distribution shift in the wild.}
Most early works on OOD detection make no distinctions on the type of distribution shift observed at test time, and create synthetic ID/OOD pairs using different datasets based on the setup in \citet{hendrycks2016baseline}.
Recently, there is an increasing interest in studying real-world distribution shifts \cite{DBLP:conf/aaai/AhmedC20,DBLP:conf/cvpr/HsuSJK20,DBLP:journals/corr/abs-1911-11132,koh2020wilds}.
On these benchmarks with a diverse set of distribution shifts,
no single detection method wins across the board.
We explore the framework of characterization of distribution shifts along the two axes of semantic shift and background (or non-semantic) shift, shedding light on the performance of current methods.

\paragraph{OOD detection in NLP.}
Even though OOD detection is crucial in production (e.g. dialogue systems \cite{ryu2018ood}) 
and high-stake applications (e.g. healthcare \cite{borjali2020natural}),
it has received relatively less attention in NLP until recently.
Recent works evaluated/improved the calibration of pretrained transformer models \cite{hendrycks2020pretrained,goyal2020evaluating,kong2020calibrated,DBLP:journals/corr/abs-2104-08812}.
They show that while pretrained transformers are better calibrated, making them better at detecting OOD data than previous models, there is scope for improvement. 
Our analysis reveals one limitation of calibration-based detection when faced with a background shift.
Other works focus on specific tasks,
including prototypical network for low-resource text classification \cite{DBLP:conf/emnlp/TanYWWPCY19}
and data augmentation for intent classification \cite{ZhengCH20}.

\paragraph{Inductive bias in OOD detection.}
Our work shows that the effectiveness of a method largely depends on whether its assumption on the distribution shift matches the test data.
One straightforward way to incorporate prior knowledge on the type of distribution shift is through augmenting similar OOD data during training, \customie, the so-called outlier exposure method \cite{DBLP:conf/iclr/HendrycksMD19},
which has been shown to be effective on question answering \cite{DBLP:conf/acl/KamathJL20}.
Given that the right type of OOD data can be difficult to obtain,
another line of work uses a hybrid of calibration and density estimation methods
to achieve a balance between capturing semantic features and background features.
These models are usually trained with both a discriminative loss and a generative (or self-supervised) loss \cite{DBLP:journals/corr/abs-2007-05566,zhang2020open,nalisnick2019hybrid}.

\textcolor{black}{\paragraph{Domain adaptation versus OOD detection.}
There are two ways of handling the effect of OOD data: 1) build models that perform well across domains (\customie, background shifts), \customie, domain adaptation \cite{DBLP:conf/coling/ChuW18, DBLP:conf/naacl/KashyapHKZ21} or 2) allow models to detect a shift in data distribution, and potentially abstain from making a prediction. In our setting (2), we want to guard against all types of OOD data without any access to it, unlike domain adaptation which usually relies on access to OOD data. This setting can be more important than (1) for safety-critical applications, such as those in healthcare, because the potential cost of an incorrect prediction is greater, motivating a more conservative approach to handling OOD data by abstaining. This could also help improve performance in selective prediction \cite{DBLP:conf/acl/KamathJL20, DBLP:conf/acl/XinTYL20}.}
\section{Conclusion}
Despite the extensive literature on outlier and OOD detection, previous work in NLP tends to lack consensus on a rigorous definition of OOD examples, instead relying on arbitrary dataset pairs from different tasks. In our work, we approach this problem in natural text and simulated data by categorizing OOD examples as either \textit{background} or \textit{semantic} shifts and study the performance of two common OOD detection methods---calibration and density estimation.
%
%
For both types of data, we find that density estimation methods outperform calibration methods under background shifts while the opposite is true under semantic shifts. However, we find several failure cases from challenge examples that target model shortcomings.

\textcolor{black}{As explained in Section \ref{sec:ood-text}, we assume that $\phi_s$ and $\phi_b$ map $x$ to two disjoint sets of components for simplicity. This assumption helps us simplify the framework and compare the two types of detection methods in relation to the two types of shifts.} While this simplified framework explains much of the differences between the two methods, failure cases from challenge examples highlight the room for better frameworks and a more explicit definition of OOD to progress the development of OOD detection methods. Such a definition can inform the creation of benchmarks on OOD detection that reflect realistic distribution shifts. 

Defining (or at least explicitly stating) the types of OOD examples that predictors are designed to target can also guide future modeling decisions between using calibration and density estimation methods, and help improve detection. Some promising directions include test-time fine-tuning \cite{DBLP:conf/icml/SunWLMEH20} and data augmentation \cite{DBLP:journals/corr/abs-2006-15207}, which can be guided towards a specific type of distribution shift for improved detection performance against it. Finally, the methods we studied work well for one type of shift, which motivates the use of hybrid models \cite{DBLP:conf/eccv/0002LG020, DBLP:journals/corr/abs-2007-09070} that use both calibration and density estimation when both types of shift occur at the same time. 





\section*{Ethical Considerations}

As society continues to rely on automated machine learning systems to make important decisions that affect human lives, OOD detection becomes increasingly vital to ensure that these systems can detect natural shifts in domain and semantics. If medical chat-bots cannot recognize that new disease variants or rare co-morbidities are OOD while diagnosing patients, they will likely provide faulty and potentially harmful recommendations \footnote{\url{https://www.nabla.com/blog/gpt-3/}} if they don't contextualize their uncertainty. We believe that implementing OOD detection, especially for more challenging but commonly occurring semantic shifts should be part of any long-lasting production model. 

In addition, OOD detection can be used to identify and alter model behavior when encountering data related to minority groups. For example, \citet{DBLP:journals/corr/abs-2012-07421} present a modified version of the CivilComments dataset \cite{10.1145/3308560.3317593}, with the task of identifying toxic user comments on online platforms. They consider domain annotations for each comment based on whether the comment mentions each of 8 demographic identities - \textit{male, female, LGBTQ, Christian, Muslim, other religions, Black and White}. They note that a standard BERT-based model trained using ERM performs poorly on the worst group, with a 34.2 \% drop in accuracy as compared to the average. Such models may lead to unintended consequences like flagging a comment as toxic just because it mentions some demographic identities, or in other words, belongs to some domains. Our work can be useful in altering the inference-time behavior of such models upon detection of such domains which constitute a larger degree of background shift. Of course, nefarious agents could use the same pipeline to alter model behavior to identify and discriminate against demographics that display such background shifts. 

\section*{Acknowledgements}
\textcolor{black}{We thank Samsung Advanced Institute of Technology (Next Generation Deep Learning: From Pattern Recognition to AI) for their support. Further, we thank the anonymous reviewers, Richard Pang, Ethan Perez, Angelica Chen and other members of the Machine Learning for Language Lab at New York University for their thoughtful suggestions on improving the paper. We also want to thank Diksha Meghwal, Vaibhav Gadodia and Ambuj Ojha for their help with an initial version of the project and experimentation setup.}

\bibliographystyle{acl_natbib}
\bibliography{anthology, custom, all}

\clearpage
\appendix
\section{Example Probability}
\label{app:prob}
We additionally evaluate our density estimation methods using $\log p(x)$ as a detection measure. In the case of text, $\log p(x)$ is defined as $\sum_{i=1}^t \log p(x_i \mid x_{<i})$.

\begin{figure}
    \centering
    \includegraphics[width=\linewidth]{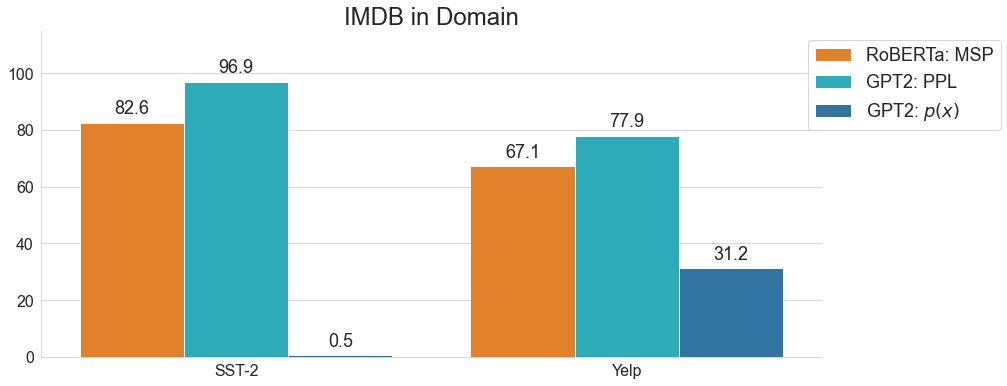}
    \caption{OOD detection performance as measured by AUROC using different measures for binary sentiment classification based background shift, using IMDB as ID data. We can see that using $\log p(x)$ as a measure is highly noisy due to its dependency on sequence lengths.}
    \label{fig:imdb_px}
\end{figure}

While \PPL~accounts for varying sequence lengths by averaging word likelihoods over the input sequence, $\log p(x)$ does not. Figure \ref{fig:imdb_px} shows that this difference significantly impacts performance. With IMDB as the ID data, using $\log p(x)$ fails for SST-2, achieving close to 100 FAR95 and near 0 AUROC. We suspect this because IMDB examples are a full paragraph while SST-2 examples are one to two sentences. $\log p(x)$ would naturally be smaller for IMDB examples than these OOD examples, resulting in complete failure for simple thresholding methods measured by AUROC.

\section{FAR95 Results}
\label{app:far95}
We additionally evaluate the performance for all experiments using FAR95, which measures the false positive rate at 95\% recall. 
In the context of OOD detection, this measure gives the misclassification rate of ID data at 95\% recall of OOD classification, hence a lower value indicates better performance.

\begin{table}
    \small
    \centering
    \setlength\tabcolsep{3pt}
    \begin{tabular}{llllr}
        \toprule
        \multirow{2}{*}{ID} &  \multirow{2}{*}{OOD} & \multicolumn{3}{c}{FAR95 ($\downarrow$)} \\
        & & \PPL & \MSP & Oracle \\
        \midrule
        News Top-5 &    News Rest & 88.5 & \textbf{75.7} &    80.4 \\
        DBPedia Top-4 & DBPedia Rest & \textbf{78.3} & 86.3 &     1.3 \\
        \bottomrule
    \end{tabular}
    \caption{FAR95 scores obtained using \PPL, \MSP~and Oracle for semantic shifts, with lower score (among \PPL/\MSP) in \textbf{bold}.}
    \label{tab:semantic-shift-far95}
\end{table}

\begin{table}
    \centering
    \small
    \begin{tabular}{ll>{\bfseries}rrr}
        \toprule
        \multirow{2}{*}{ID} &  \multirow{2}{*}{OOD} & \multicolumn{3}{c}{FAR95 ($\downarrow$)} \\
        & & \textnormal{\PPL} & \MSP & Oracle \\
        \midrule
        \multirow{2}{*}{SST-2} &       IMDB &  8.6 & 76.5 &     0.0 \\
         &       Yelp &  5.2 & 83.0 &     0.0 \\
        \midrule
        \multirow{2}{*}{IMDB} &      SST-2 & 17.0 & 47.7 &     0.2 \\
         &       Yelp & 70.2 & 82.6 &     0.0 \\
        \midrule
        \multirow{2}{*}{Yelp} &      SST-2 &  3.1 & 45.4 &     1.1 \\
         &       IMDB & 36.2 & 90.4 &     0.0 \\
        \midrule
        \multirow{2}{*}{SNLI} 
             &       RTE & 19.1 & 61.4 &   0.7 \\
             &       MNLI & 14.7 & 62.5 &    0.3 \\
        \midrule
        \multirow{2}{*}{RTE} & SNLI & 62.5 & 95.3 &   0.0 \\
            &       MNLI & 64.3 & 93.9 &   10.3 \\
        \midrule
        \multirow{2}{*}{MNLI} &      SNLI & 70.9 & 84.6 &    1.2 \\
            &       RTE & \textnormal{93.2} & \textbf{69.8} &   6.2 \\
        \bottomrule
    \end{tabular}
    \caption{FAR95 scores obtained using \PPL, \MSP~and Oracle for background shift caused by shift in domain. For each pair, lower score obtained (by \PPL~or \MSP) is in \textbf{bold}.}
    \label{tab:domain-shift-far95}
\end{table}

\begin{table}
    \centering
    \small
    \resizebox{\columnwidth}{!}{%
    \begin{tabular}{lr>{\bfseries}rrrr}
        \toprule
        \multirow{2}{*}{ID} &  \multirow{2}{*}{OOD} & \multicolumn{3}{c}{FAR95 ($\downarrow$)} \\
        & & \textnormal{\PPL} & \MSP & Oracle \\
        \midrule
        \multirow{4}{*}{Fiction} & Government & 57.4 & 95.0 &     9.7 \\
         &      Slate & 66.0 & 92.7 &    37.7 \\
         &  Telephone & 29.1 & 93.3 &    36.0 \\
         &     Travel & 58.0 & 93.3 &    10.0 \\
        \midrule
        \multirow{4}{*}{Government} &    Fiction & 74.7 & 92.6 &     6.4 \\
         &      Slate & 70.7 & 92.1 &    13.7 \\
         &  Telephone & 35.2 & 95.5 &     6.2 \\
         &     Travel & 52.8 & 92.4 &     6.2 \\
        \midrule
        \multirow{4}{*}{Slate} &    Fiction & 90.6 & 96.2 &    32.2 \\
         & Government & 90.0 & 96.1 &    12.6 \\
         &  Telephone & 57.4 & 96.0 &    22.7 \\
         &     Travel & 83.3 & 95.8 &    16.8 \\
        \midrule
        \multirow{4}{*}{Telephone} &    Fiction & 54.2 & 93.3 &    32.5 \\
         & Government & 50.9 & 93.7 &     8.5 \\
         &      Slate & 49.6 & 91.1 &    36.3 \\
         &     Travel & 44.6 & 91.4 &    10.7 \\
        \midrule
        \multirow{4}{*}{Travel} &    Fiction & 74.5 & 95.5 &    10.2 \\
         & Government & 69.0 & 94.4 &     7.8 \\
         &      Slate & 75.9 & 93.8 &    16.8 \\
         &  Telephone & 30.3 & 93.7 &     9.5 \\

        \bottomrule
    \end{tabular}
    }
    \caption{FAR95 scores obtained using \PPL, \MSP~and Oracle for background shift caused by shift in MNLI genre. For each pair, lower score obtained (by \PPL~or \MSP) is in \textbf{bold}.}
    \label{tab:MNLI-genre-far95}
\end{table}

Tables \ref{tab:semantic-shift-far95}, \ref{tab:domain-shift-far95}, \ref{tab:MNLI-genre-far95} and \ref{tab:MNLI-challenge-far95} show the results obtained using FAR95 as a metric for the corresponding ID/OOD pairs used earlier. We observe that FAR95 results are in line with AUROC results except for DBPedia, in which case density estimation methods yield a better result. The difference may be a result of the accumulative nature of AUROC in contrast to FAR95, which is a point measurement.

\begin{table}
    \small
    \centering
    \setlength\tabcolsep{3pt}
    \resizebox{\columnwidth}{!}{%
    \begin{tabular}{lllr>{\bfseries}rr}
    \toprule
    \multirow{2}{*}{ID} &  \multirow{2}{*}{OOD} & \multirow{2}{*}{Shift} & \multicolumn{3}{c}{FAR95 ($\downarrow$)} \\
        & & & \PPL & \textnormal{\MSP} & Oracle \\
    \midrule
    IMDB &       c-IMDB & Semantic & 93.1 & 82.8 &    \textbf{69.3} \\
    \midrule
    \multirow{7}{*}{MNLI} &    HANS & Background &  \textbf{4.2} & \textnormal{73.1} &     0.0 \\
      &            Negation & Background &  94.9 & 93.5 &     0.1 \\
      &     Len. Mismatch & Background &  98.3 & 95.0 &     0.1 \\
      &      Spell. Error & Background &  96.9 & 92.4 &     3.0 \\
      &        Word Overlap & Background &  96.0 & 94.4 &     1.1 \\
      &             Antonym & Semantic & 100.0 & 90.8 &     6.3 \\
      & Num. Reas. & Semantic &  99.5 & 77.6 &     0.7 \\
    \bottomrule
    \end{tabular}
    }
    \caption{FAR95 scores obtained using \PPL, \MSP~and Oracle for challenge data. The primary type of shift observed is indicated in the 'Shift' column. Lower score (among \MSP/\PPL) for each pair is in \textbf{bold}.}
    \label{tab:MNLI-challenge-far95}
\end{table}

\section{Background shift in MNLI Genres}
MNLI is a crowd-sourced collection of sentence pairs for textual entailment sourced from 10 genres including Fiction, Government, Slate, Telephone, and Travel. We use examples from these five MNLI genres and separately consider each genre as ID and OOD, using the validation splits for evaluation.

Table \ref{tab:MNLI-genre} shows the results for MNLI genres. The discriminative model generalizes well to other genres and we find that the OOD detection performance of calibration method is close to random (50) because of the higher confidence on correct OOD predictions by a well-calibrated model.

\begin{table}
    \centering
    \small
    \resizebox{\columnwidth}{!}{%
    \begin{tabular}{lr>{\bfseries}rrrrr}
        \toprule
        \multirow{2}{*}{ID} &  \multirow{2}{*}{OOD} & \multicolumn{3}{c}{AUROC} & \multicolumn{2}{c}{Accuracy} \\
        & & \textnormal{\PPL} & \MSP & Oracle & OOD & ID \\
        \midrule
        \multirow{4}{*}{Fiction} & Govt. & 83.3 & 48.5 &    98.4 & 87.0 & \multirow{4}{*}{86.1} \\
         &      Slate & 81.6 & 54.1 &    92.7 & 82.2 \\
         &  Tel. & 92.3 & 51.0 &    94.6 & 84.0 \\
         &     Travel & 82.2 & 49.9 &    98.3 & 84.3 \\
         \midrule
        \multirow{4}{*}{Govt.} &    Fiction & 75.2 & 57.4 &    98.9 & 82.8 & \multirow{4}{*}{88.4} \\
         &      Slate & 77.1 & 58.3 &    97.7 & 82.0 \\
         &  Tel. & 89.9 & 57.6 &    98.5 & 82.8 \\
         &     Travel & 82.6 & 57.1 &    99.4 & 84.1 \\
         \midrule
        \multirow{4}{*}{Slate} &    Fiction & 60.6 & 48.2 &    94.2 & 84.3 & \multirow{4}{*}{82.5} \\
         & Govt. & 61.3 & 45.3 &    97.7 & 87.8 \\
         &  Tel. & 83.0 & 49.8 &    95.3 & 84.1 \\
         &     Travel & 63.7 & 46.8 &    97.6 & 84.6\\
         \midrule
        \multirow{4}{*}{Tel.} &    Fiction & 85.7 & 55.9 &    95.2 & 82.5 & \multirow{4}{*}{85.7} \\
         & Govt. & 86.0 & 52.5 &    98.3 & 85.9 \\
         &      Slate & 86.8 & 59.2 &    94.2 & 80.6 \\
         &     Travel & 87.8 & 56.8 &    98.6 & 82.5 \\
         \midrule
        \multirow{4}{*}{Travel} &    Fiction & 76.4 & 54.8 &    98.0 & 81.3 & \multirow{4}{*}{86.7} \\
         & Govt. & 78.8 & 49.0 &    98.7 & 87.4 \\
         &      Slate & 77.2 & 55.8 &    96.3 & 80.8 \\
         &  Tel. & 92.7 & 56.0 &    98.1 & 82.2 \\
        \bottomrule
    \end{tabular}
    }
    \caption{Performance on background shifts caused by shift in MNLI genre. For each pair, higher score obtained (by \PPL~or \MSP) is in \textbf{bold}. We can see that the density estimation method using \PPL~significantly outperforms the calibration method.}
    \label{tab:MNLI-genre}
\end{table}

\end{document}